\documentclass[runningheads]{llncs}
\input epsf
\usepackage{graphicx}
\usepackage{algorithm}

\begin{document}
\title{Exploring Graph Representation of Chorales} 
\titlerunning{Graph Representation}
\author{Somnuk Phon-Amnuaisuk\inst{1}\orcidID{0000-0003-2130-185x} } 
\institute{Media Informatics Special Interest Group,\\
Universiti Teknologi Brunei, Gadong, Brunei \\
\email{somnuk.phonamnuaisuk@utb.edu.bn}
}

\maketitle   
\begin{abstract}
This work explores areas overlapping music, graph theory, and machine learning. An embedding representation of a node, in a weighted undirected graph $\mathcal{G}$, is a representation that captures the meaning of nodes in an embedding space. In this work, 383 Bach chorales were compiled and represented as a graph. Two application cases were investigated in this paper (i) learning node embedding representation using \emph{Continuous Bag of Words (CBOW), skip-gram}, and \emph{node2vec} algorithms, and (ii) learning node labels from neighboring nodes based on a collective classification approach. The results of this exploratory study ascertains many salient features of the graph-based representation approach applicable to music applications.

\keywords{Graph representation  \and Chorales \and Learning node embedding \and Node2Vec \and Collective classification }
\end{abstract}

\section{Introduction}
The \emph{node2vec} method \cite{grover16} is applied to learn low-dimensional embedding representation of nodes in a chorale graph. The attempts at a novel application of graph representation learning in the chorales domain entails many challenges. Hence, methodology and appropriate data must be prepared from scratch. Thanks to Music21 \cite{cuthbert10}, the process of chorales data preparation has been simplified substantially. We prepare harmonic progressions from three hundred and eighty-three (383) chorales retrieved from the Bach chorales corpus available in Music21. A chorale graph is created based on similarities between harmonic sequences of chorales. A low dimensional-embedding representation that preserves the original neighborhood structure of the graph is learned using the following three techniques: continuous bag of words (\emph{CBOW}), \emph{skip-gram}, and \emph{node2vec}. Both CBOW and skip-gram are available from the \emph{word2vec} \cite{w2v13} class in \emph{Gensim} library \cite{radimrehurek10}).

Two application cases were presented in this paper (i) learning node embedding representation and then suggesting similar chorales based on similarities in the embedding space, and (ii) learning node labels from neighboring nodes using the collective classification approach. The rest of the paper is organized as follows: Section 2 discusses the background and some related works; Section 3 discusses our approach and gives the details of the techniques behind it; Section 4 provides a critical evaluation and discussion of the empirical results; and the final section presents the conclusion and further research.

\section{Graph Representation Learning of Chorales}

Two important criteria in music representation systems: \emph{expressive completeness} and \emph{structural generality} were discussed in \cite{awh93}. Expressive completeness refers to the range of raw musical data that can be represented, and structural generality refers to the range of higher-level structures that can be represented and manipulated using the representation system.

Since traditional music notations have been developed to abstract sound streams, the representation system for the purpose of music analysis commonly represents the basic building blocks of pitches, durations and facilitates functionalities on them for the derivation of more complex knowledge from these building blocks \cite{RPI91,courtot92,span06}.
For example:

\begin{center}
\begin{small}
\begin{tabular}{|c|rl|}   \hline
&Degree : & \{$1,2,3,4,5,6,7$\} \\
 Basic types &Accidental : &  \{$\natural,\sharp,\times,\flat,\flat\flat$\} \\
&Octave : &  \{1..8\} \\ 
&Pitch : &  $\langle$ Degree, Accidental, Octave $\rangle$ \\ \hline

& $add_{dd}$  :& Duration $\times$ Duration $\rightarrow$ Duration\\
 Basic operations & $add_{td}$  :& Time $\times$ Duration $\rightarrow$ Time\\
& $sub_{tt}$  :& Time $\times$ Time $\rightarrow$ Duration\\
& $sub_{dd}$  :& Duration $\times$ Duration $\rightarrow$ Duration\\  \hline 
\end{tabular}
\end{small}
\end{center}
In this work, we resort to Music21, a python-based toolkit that facilitates knowledge representation and scores well in both expressive completeness, and structural generality criteria discussed above. 

\subsection{Representing Music as a Graph}
A musical dice game \emph{(Musikalisches W\"urfelspiel)} is one of the formal compositional processes invented since the 1700s \cite{cope01} (page 4). This process can be formulated as a multi-graph where each node represents a measure (i.e. bar) of pre-composed materials linking to various nodes. This graph can be employed to generate a new piece of music by traversing the graph in accordance with imposed constraints. 

Describing domain knowledge using graphs by abstracting knowledge into nodes,  and edges (e.g., social networks, communication networks, molecule or protein structures, visual scene graph) offers additional benefits from information inherited or derived from its structures as they present information which would otherwise be unavailable in traditional non-graph representation approaches.

In \cite{orio09}, a graph structure is employed to represent music scores where terminal nodes describe melodic contents, the internal node represents its incremental generalization and the edge represents a relationship between nodes. The graph structure in \cite{orio09} is employed to calculate melodic similarities for content-based music retrieval tasks. The design of the graph can be in various forms e.g., monopartite graphs (single class of nodes), multipartite graphs, simple graphs, and multigraphs. The design is application-dependent. In \cite{jeong19}, a graph neural network is designed for music score data and modeling expressive piano performance.

\section{Materials and Methods}
Music21 provides various modules for musicologists to analyze music. \emph{Chordify} is one of the modules provided in Music21. Chordify analyzes polyphonic notes and reduce them to a chord. These chords can be notated using a Roman numeral in a functional harmony fashion. For example, $I$ and $I6$ denotes the tonic chord in its root and first inversion, respectively. $V7$ denotes the dominant seventh chord and $viio$ denotes the diminished leading chord, respectively.  

\begin{figure}[!ht]
\begin{center}\leavevmode
\epsfxsize=12cm
\epsfbox{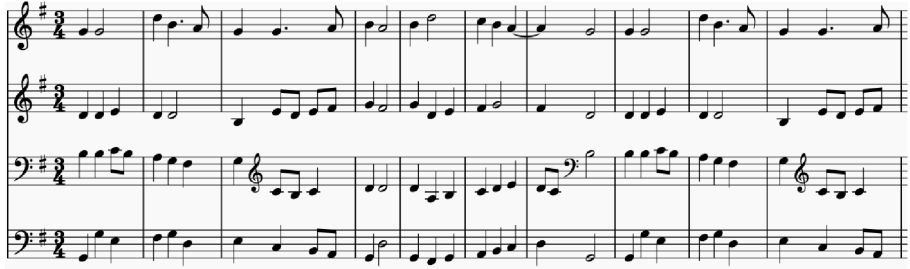}
\end{center}
\caption{The first ten measures (bars) from chorale BWV269.}
\label{bwv269}
\end{figure}

We retrieved 383 chorales from the  core corpus available in Music21\footnote{At presence, there are 387 unique Bach chorales in the Music21 corpus, we only use 383 chorales in this work since some chorales formats are not compatible with our conversion script.} and wrote a Python script to extract four voices (i.e., soprano, alto, tenor and bass) then infer chordal sequences of the four part chorales. Figure \ref{bwv269} shows the first ten (10) measures from chorale BWV269. Table \ref{harmony} illustrates representative examples of harmonic sequences extracted from chorales ID  BWV86.6,BWV148.6, BWV178.7, BWV185.6, BWV269, and BWV355 respectively.

\begin{table} 
\caption{Examples of harmonic sequences extracted from Chorales}
\begin{small}
\begin{tabular}{|c|c|c|}
\hline 
{\bf BWV} & {\bf Harmonic sequence} &  {\bf Cadence} \\ 
\hline 

86.6 & [I, V6, iii43, iv74, V7, I, bVII6, iii$\phi$7b53, I, ...] & [I, V, v5, II6, V, I] \\ 

148.6 & [i, ii743, i6, ii, viio\#63, i, i\#7, iv6, iio64, ...]  & [iv6, i64, ii$\phi$65, V, V75\#3, I] \\ 

178.7 & [i, III6, v7, bVI, iio64, III6, bVII64, III, i, ...] & [iv6, i43, ii$\phi$65, i54, V75\#3, I]  \\ 

185.6 & [V6, V65, i532, i, ii754, V75\#3, i, ii42, i, i, ...] & [iio6, III7, i752, i54, V75\#3, I] \\ 

269 & [I, I, IV6, vi, V6, I, iii6, V, vi, IV, I752, ...]  & [vi, vi42, ii65, V, V7, I] \\ 

355 & [I, I6, ii65, V, I, I42, \#ivo6, V, V, V42, I6, ...]  & [I6, ii65, ii7, V, V7, I] \\ 
\hline 
\end{tabular} 
\end{small}
\label{harmony}
\end{table}

\subsection{Compute Harmonic Word Embedding Representation}
Recent progress in natural language processing (NLP) promotes the concept of word embedding  representation which is a distributed vectorized representation of a word $w$, $f(w) \mapsto \mathcal{R}^d$.  Word embedding representation could encode the meanings of words where words that are close together in the representation space also share similar meanings or related concepts. For example, the word \emph{ice} should be closer to the word \emph{cold} rather than the word \emph{hot}.

Music chords abstract polyphonic sound into symbols (i.e., the extracted harmonic sequences of chorales). These symbols can be considered as a set of words describing the harmonic vocabulary of chorales.

Here, the \emph{word2vec} model in Gensim \cite{radimrehurek10} was employed to learn the embedding representation of chords (i.e., words) in chorales. The embedding representations were used to compute similarities between chorale nodes when constructing a chorale graph.

\subsection{Constructing Chorale Graphs}
\label{choralegraph}
Let a graph $\mathcal{G} = (\mathcal{V},\mathcal{E})$, $\mathcal{V}$ denotes a set of vertices and each vertex (node) represent a chorale composition; and $\mathcal{E}$ denotes a set of un-ordered pairs $(u,v) \in \{\mathcal{V} \times \mathcal{V}\}$. A graph  $\mathcal{G}$ is constructed by connecting any two chorales that satisfies the required constraints  $\xi$ imposed by a similarity function $\mathcal{S}$. e.g., $\mathcal{S}$ $(u,v) > \xi$.

In this paper, a graph was constructed based on the similarity of the harmonic progression in the intro, and the final cadence between two chorales. The similarity $\mathcal{S}$ between two harmonic progressions (chord sequences taken from chorales $u$ and $v$) were estimated using the Algorithm \ref{alg1}. 

\subsection{Learning Node Embedding of Chorales}
Word embedding representation from the \emph{word2vec} model is a latent variable learned by training a shallow feed-forward neural network to learn relationships between target and context words. Training samples for \emph{word2vec} model are prepared from the desired text corpus. The same tactic used in \emph{word2vec} to learn word embedding representation can be applied to learn relationships among nodes in a graph. The words are replaced with nodes, and training samples are prepared by sampling nodes from the graph.

Learning node embedding yields a compact representation of nodes in a graph. A node embedding representation $f:u \mapsto \mathcal{R}^d$ maps nodes $u$ and $v$ in a graph to vectors $f(u) \rightarrow {\bf z}_u$ and $f(v) \rightarrow {\bf z}_v$ where  ${\bf z}_u, {\bf z}_v$ still preserves the original topological structure of graph $\mathcal{G}$ in the embedding space. The more compact representation in the embedding space is effective in computing similarity using ${\bf z}_u$ and ${\bf z}_v$.
The function $f$ is unknown but can be approximated by maximizing the log likelihood of observing neighbouring node of $u$ given the embedding ${\bf z}_u$. This can be written down as the objective function below:
\begin{equation}
{max}_{f} \sum_{u \in \mathcal{V}} log P(N_r(u)| {\bf z}_u)
\end{equation} 
where $N_r(u)$ denotes a neighbourhood function.  Let us define neighbouring nodes of $u$ as nodes in the path walking from $u$ using a random walk policy $r$.

\begin{algorithm}
\caption{ Graph construction and Similarity computation}
\begin{tabbing}[!htb]
\hspace{0.8cm}  \= \hspace{0.8cm}  \= \hspace{0.80cm}  \= \hspace{0.80cm} \= \hspace{0.80cm}   \kill
\>\>\>\>\\
\> Let ${Comp}$ be a set of chorale compositions. \>\>\>\\
\> Let $Seq_u$ and $Seq_v$ be any two chord sequences having the same length $n$.\>\>\>\\
\> Let $c_i, c_j$ be chords in the sequence and $i, j$ denote index $1 ... n$. \>\>\>\\
\> Let $w2v(c_i,c_j) \mapsto \mathcal{R}$ be a function that computes similarity between chords.\>\>\>\\
\> The similarity value is weighted by attention score $e^{-|i-j|}$ emphasizing  \>\>\>\\
\> a chord pair with the same position $f(w) \mapsto \mathcal{R}^d$. index.\>\>\>\\
\>\>\>\>\\
\> {\bf procedure} $GraphConstruction$ $(Comp)$ \>\>\>\\
\>\>${\bf forall }$ $(u,v) \in \{Comp \times Comp\}$ and $u \neq v$  \>\>\\
\>\>\>{\bf if} $\mathcal{S}$ $(u,v)$ complies to constraints $\xi$ \>\\
\>\>\>{\bf then} add nodes $u$, $v$ to graph $\mathcal{G}$ \>\\
\>\>\>\> and add edge $(u,v)$ with weight $\mathcal{S}$ $(u,v)$  \\
\>\>{\bf return} $\mathcal{G}$ \> \>\\
\>\>\>\>\\
\> {\bf procedure} $\mathcal{S}$ $(u,v)$ \>\>\>\\
\>\> extract $Seg_u, Seq_v$ from composition $u$ and $v$  \>\>\\
\>\>$similarity$ = 0 \>\>\\
\>\>{\bf for} $c_i$ in $Seq_u$ \>\>\\
\>\>\> {\bf for} $c_j$ in $Seq_v$ \>\\
\>\>\> \> $similarity$ += $ w2v(c_i, c_j)*e^{-|i-j|}$ \\ 
\>\>\> {\bf end for} \> \\
\>\>{\bf end for} \> \>\\
\>\>{\bf return} $similarity$ \> \>\\
\end{tabbing}
\label{alg1}
\end{algorithm}

\subsubsection{Collect Training Samples using Biased Random Walk}
Starting from node $u$, all nodes in the same path starting from $u$ were considered as sharing some similarities to $u$ since it was within a predefined walking steps. Readers may foresee that the graph could be traversed in many fashions, by exploring neighboring nodes or wandering deeper and further away from the starting points. Here, the  bias random walk  was employed with two hyper-parameters: return parameter ($p$) and in-out parameter ($q$), follows \cite{grover16}. 

In brief, if the current node was $v$, all edges to all neighboring nodes $N(v)$ were weighted according to  parameters $p$ and $q$. Let $u$ be the previous node, and $w$ be the next possible node, where $u \in N(v) \wedge w \in N(v)$. The edge $(u,v)$ was assigned with the weight of 1, the edges $(v,w)$ were assigned with the weight $1/p$ if $w \in N(u)$ else the weight $1/q$ was assigned to the edges. Finally edge weights were normalized, and the values conditioned their chance of being sampled as the next node.

\begin{figure}[!ht]
\begin{center}\leavevmode
\epsfxsize=6.5cm
\epsfbox{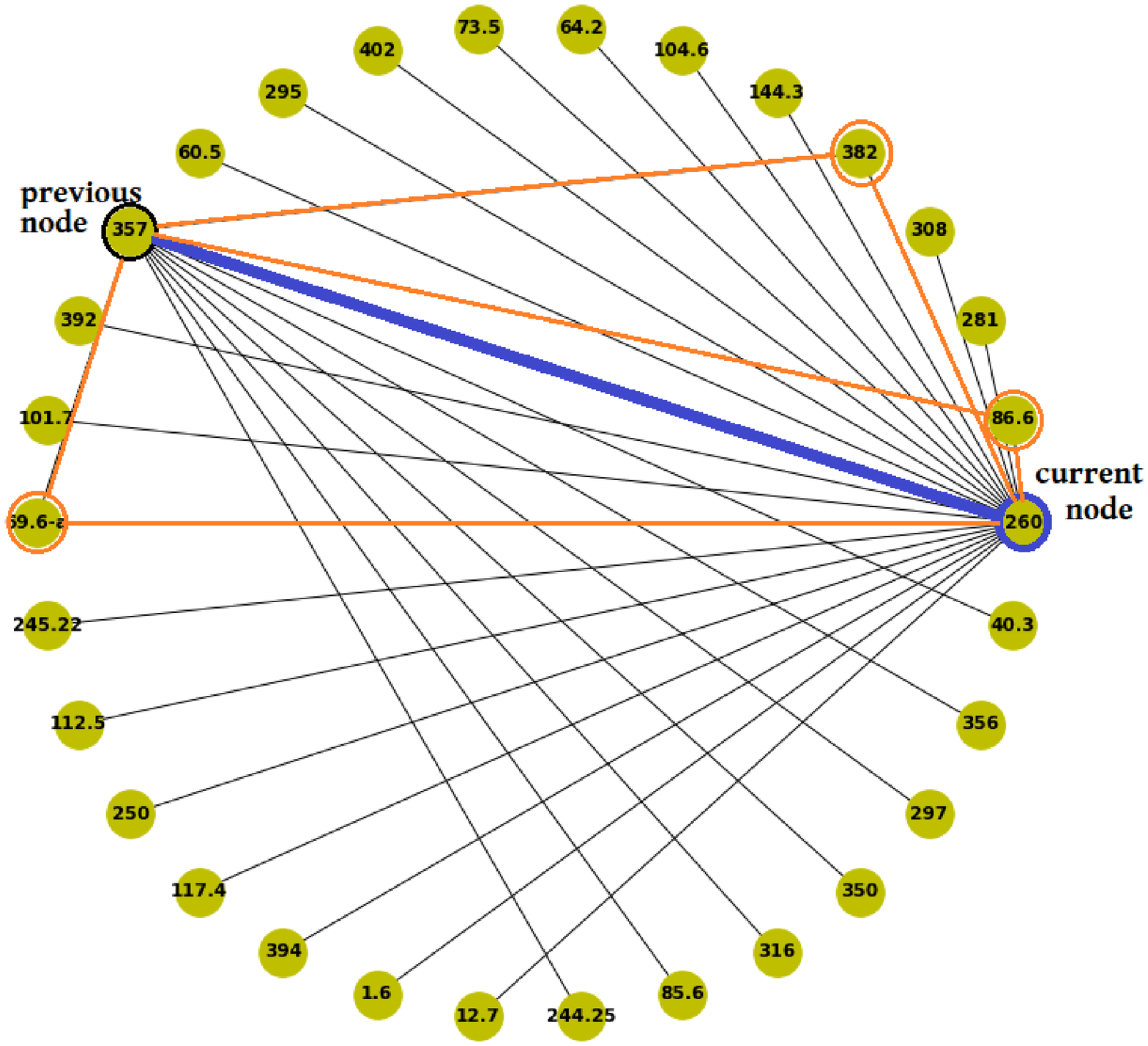}
\epsfxsize=5.5cm
\epsfbox{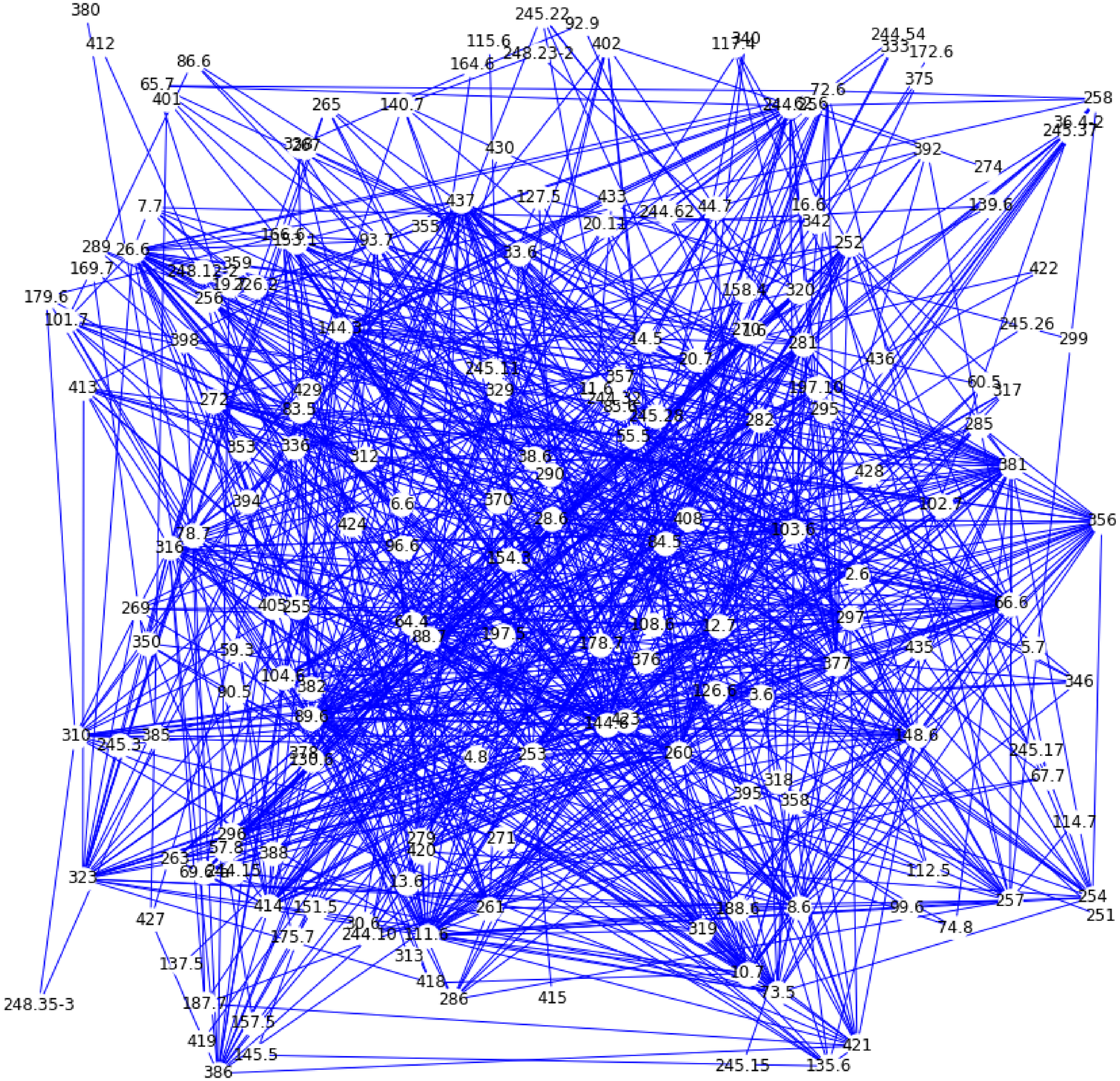}
\end{center}
\caption{Left pane: node 260 is the current node and all nodes connected to it are possible next nodes $n \in N(260)$. The edges (260, $n$) are assigned with the weight either 1,  $1/p$, or  $1/q$ (see text for detailed explanation). Right pane: an instance of a chorale graph used in our study.}
\label{neighbor1}
\end{figure}

Setting $p < 1$ encourages local exploration while setting $q < 1$ encourages global exploration since a lower value of $p$ will increase probability of exploring nodes neighboring both previous node $u$ and the current node $v$, while setting $q < 1$ encourages the next node to be further away from $u$ (see Figure \ref{neighbor1}).

One could collect nodes from many traversed paths starting from each node in the graph. Walk data collection was controlled by the number of steps in each walk and the number of repeated walks for each node. These walk data samples constituted a dataset for training \emph{node2vec} models using CBOW and skip-gram approaches. These walks were also used to prepare negative sampling training data. In a negative sampling approach \cite{w2v13}, any node pairs in the same path would be collected as positive (target, context) examples and node pairs not from the same path would be collected as negative examples.

\begin{figure}[!ht]
\begin{center}\leavevmode
\epsfxsize=10.cm
\epsfbox{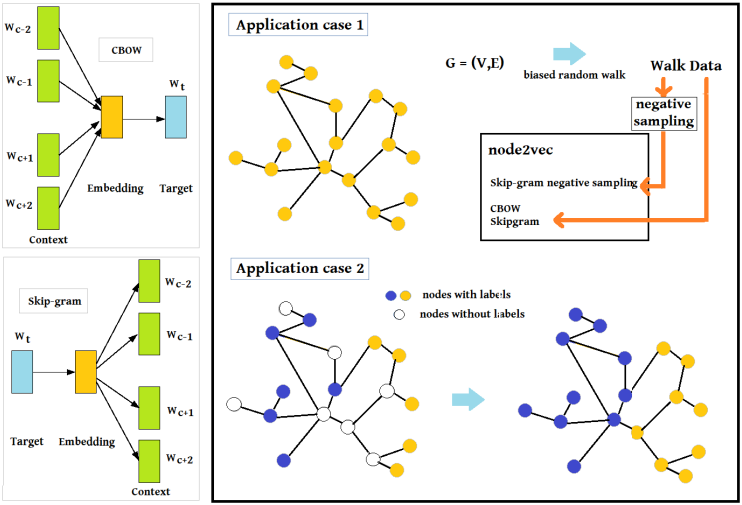}
\end{center}
\caption{Left column: architecture of CBOW and skip-gram; Right column: a graphical summary of application case 1 and 2.}
\label{node2vecmodels}
\end{figure}

\section{Empirical Results}
Two application cases are discussed in this section: (i) learning node embedding representation and (ii) learning node labels from neighbouring nodes. Both applications are based on the following assumptions, given a graph $\mathcal{G} = (\mathcal{V,E})$: (i) $similarity(u,v) \approx {\bf z}_u {\bf z}_v$,   for $u,v \in \mathcal{V}$; (ii) Node without label can be inferred from labels from neighbouring nodes. Figure \ref{node2vecmodels} provides a graphical summary of our empirical study.

\subsection{Application Case 1: Learning Node Embedding Representation}
A chorale graph, consisting of 194 nodes 861 edges, and average node degree of 8.88, was created using similarities among the first six chords (as discussed in section \ref{choralegraph}). The construction of this graph can be conceptually understood as, starting with a complete graph, then removing edges that do not fulfil specified constraints from the chorale graph, and finally remove all isolated nodes.

Three node embedding algorithms were explored, the first model was the \emph{node2vec} trained using  the negative sampling tactic \cite{grover16}. Negative sampling is a form of noise contrastive estimation (NCE) \cite{goldberg14} that approximates the log probability of $P(w|{\bf z}_v)$. Instead of normalizing with all nodes in the graph i.e., $\sum_{w \in N_r(v)}  log(P(w|{\bf z}_v))$, it is normalized using only $k$ random negative samples  i.e., $ \sum_{i=1}^k log( \sigma({\bf z}_v^T,{\bf z}_{n_i}),$ where $\sigma$ denotes a sigmoid function). The negative ${\bf z}_{n_i}$ examples are randomly sampling over relevant nodes in the graph with a bias towards nodes with higher degree values.
\begin{algorithm} 
\caption{Learning node embedding using \emph{node2vec\_SGNS}, SG and CBOW}
\begin{tabbing}
\hspace{0.1cm}  \= \hspace{0.4cm}  \= \hspace{0.4cm}  \= \hspace{0.4cm} \= \hspace{0.4cm}   \kill
\>\>\>\>\\
\> Let ${Comp}$ be a set of chorale compositions. \>\>\>\\
\> Let $\mathcal{G} = (\mathcal{V}, \mathcal{E})$, $\mathcal{V} \subseteq Comp$ and $\mathcal{E} \subseteq \{\mathcal{V} \times \mathcal{V}\}$. \>\>\>\\
\> Let $\tau_u$ be  an $n$-step random walk traversal starting from node $u$.\>\>\>\\
\>\>\>\>\\
\> {\bf node2vec with skip-gram negative sampling} \>\>\>\\
\> Let $(u,w,lab)$ denotes target $u$ and context $w$. If $lab=1$ (positive example) \>\>\>\\
\> then $u \in \tau_u$ and  $w \in \tau_u$, if $lab=0$ (negative example) then  $u \in \tau_u$ and  $w  \notin \tau_u$.\>\>\>\\
\> Let $D$ be the training data composed of positive and negative samples. \>\>\>\\
\> Let ${\bf Z} \in \mathcal{R}^{d \times |\mathcal{V}|}$ denotes an encoder matrix. \>\>\>\\
\> Let $u,w \in \mathcal{I}^{|\mathcal{V}|}$ denotes one hot encoding of node $u$ and $w$, respectively. \>\>\>\\
\>\>\>\>\\
\> {\bf procedure} $node2vec\_SGNS$ $(D)$ \>\>\>\\
\>\> $initialize( {\bf Z} )$ to random values ${\bf Z} \in \mathcal{R}^{d \times |\mathcal{V}|}$ \>\>\\
\>\> ${\bf repeat}$ epochs \>\>\\
\>\>\> ${\bf forall}$ $(u, w, lab) \in D$ \>\\
\>\>\>\> ${\bf z}_u = {\bf Z} \cdot u$ \\
\>\>\>\> ${\bf z}_w = {\bf Z} \cdot w$ \\
\>\>\>\> $P(w|{\bf z}_u) = \frac{exp({\bf z}_u^T{\bf z_w})}{\sum_{n \in \mathcal{V}} exp({\bf z}_u^T{\bf z_n})}$ \\
\>\>\>\> $\mathcal{L} = - \sum_{u \in \mathcal{V}} \sum_{w \in N_r(u)}  log(P(w|{\bf z}_u)$ \\ 
\>\>\>\> \hspace{0.5cm} $\approx -  \sum_{u \in \mathcal{V}} \sum_{w \in N_r(u)}
[ log ( \sigma({\bf z}_u^T,{\bf z}_w) ) - \sum_{i=1}^k log( \sigma({\bf z}_u^T,{\bf z}_{n_i})) ]$ \\ 
\>\>\>\> ${\bf z}_w \leftarrow {\bf z}_w - \eta \frac{\partial \mathcal{L}^{(u)}}{\partial {\bf z}_w}$ \\ 
\>\>\>\>\\
\> {\bf Learning node embedding with CBOW, and skip-gram} \>\>\>\\
\> Let $D$ be the training data composed of walk $\tau_u, \forall u \in \mathcal{V}$ \>\>\>\\
\> Let $c \in C$ be node in the walk $\tau_u$, $C$ denotes context nodes $C \subseteq \tau_u$  \>\>\>\\
\>\>\>\>\\
\> {\bf procedure} $CBOW\_SG$ $(D, method)$ \>\>\>\\
\>\> $initialize( {\bf Z} )$ to random values ${\bf Z} \in \mathcal{R}^{d \times |\mathcal{V}|}$ \>\>\\
\>\> ${\bf repeat}$ epochs \>\>\\
\>\>\>  ${\bf forall}$  $u \in \tau_u, C \subseteq \tau_u$ and $\tau_u \in D$ \>\\
\>\>\>\> {\bf if} method {\bf is} CBOW \\
\>\>\>\> \hspace{0.5cm}Given $P(u|C,{\bf Z}^C)$ where the expectation ${\bf Z} = \frac{1}{|C|}{\bf Z}^C_c$ $\forall$ $c \in C$ \\
\>\>\>\> \hspace{0.5cm}$P(u|c,{\bf Z}) = \frac{exp({\bf z}_c^T{\bf z}_u)}{\sum_{n \in V} exp({\bf z}_c^T{\bf z}_n)}$ \\
\>\>\>\> \hspace{0.5cm}$\mathcal{L} = - \sum_{c \in C}  \sum_{u \in \mathcal{V}}  log(P(u|c,{\bf Z})$ \\ 
\>\>\>\> {\bf if} method {\bf is}  skip-gram \\
\>\>\>\> \hspace{0.5cm}$P(C|u,{\bf Z}) = \prod_{c \in C} P(c|u,{\bf Z})$ \\
\>\>\>\> \hspace{0.5cm}$P(c|u,{\bf Z}) = \frac{exp({\bf z}_u^T{\bf z}_c)}{\sum_{n \in V} exp({\bf z}_u^T{\bf z}_n)}$ \\
\>\>\>\> \hspace{0.5cm}$\mathcal{L} = -  \sum_{c \in C} \sum_{u \in \mathcal{V}}  log(P(c|u,{\bf Z})$ \\ 
\end{tabbing}
\label{alg2}
\end{algorithm}

The other two node embedding models were trained using two popular approaches CBOW and skip-gram from \emph{word2vec} class \cite{w2v13} provided in the Gensim \cite{radimrehurek10}. Algorithm \ref{alg2} outlines the concepts of the methods used in this work. 

\begin{table} 
\caption{Three examples of similar chorales suggested by the models, BWV148.6-BWV316, BWV253-BWV414, and BWV318-BWV355.}
\begin{center}
\begin{small}
\begin{tabular}{|c|c|}
\hline 
{\bf BWV} & {\bf Excerpts from chorales}  \\ 
\hline 
148.6 & \\
& \epsfxsize=10.cm \epsfbox{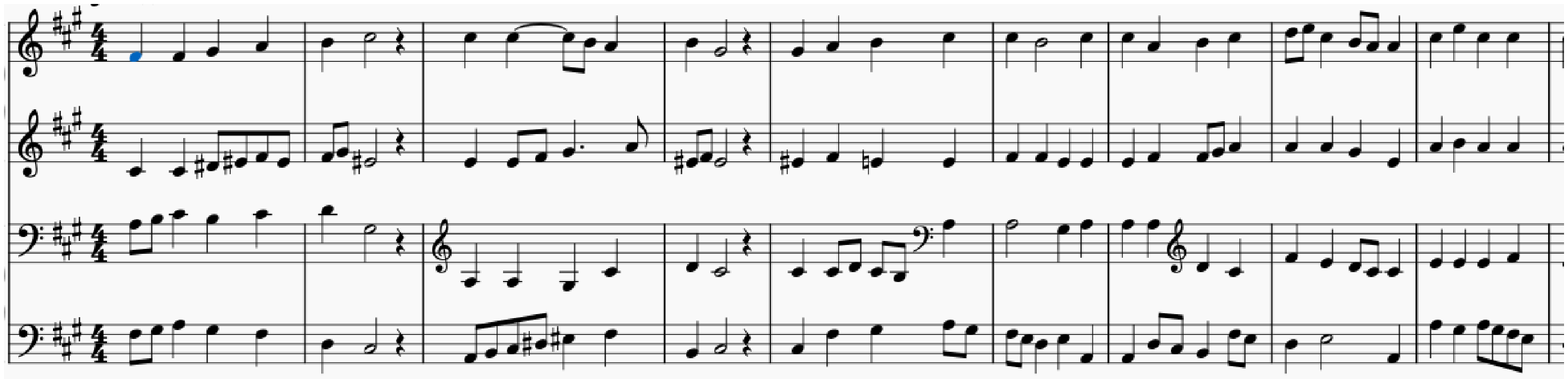}  \\  
316 & \\
&  \epsfxsize=10.cm \epsfbox{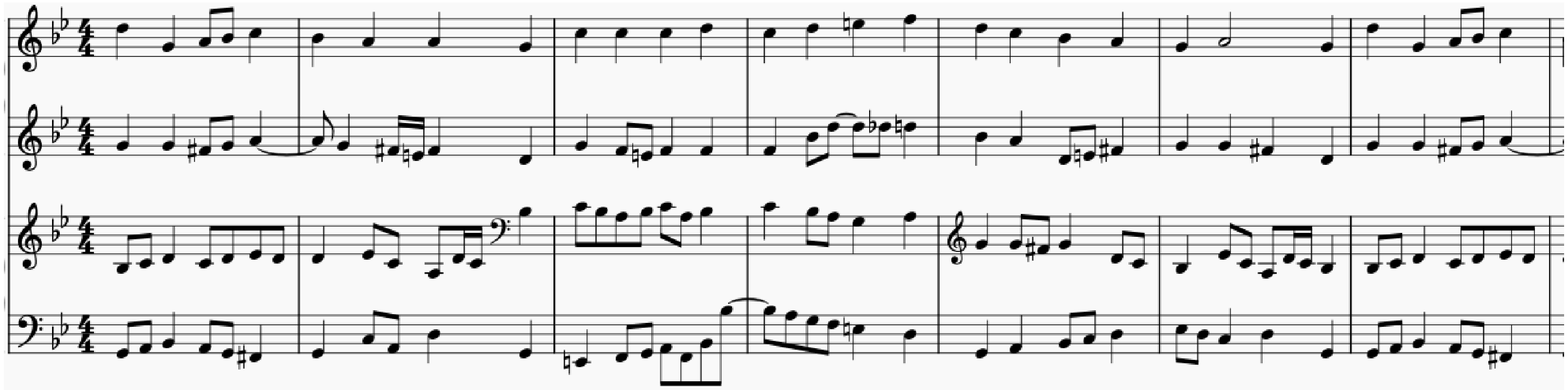}  \\  \hline \hline
253 &\\
& \epsfxsize=10.cm \epsfbox{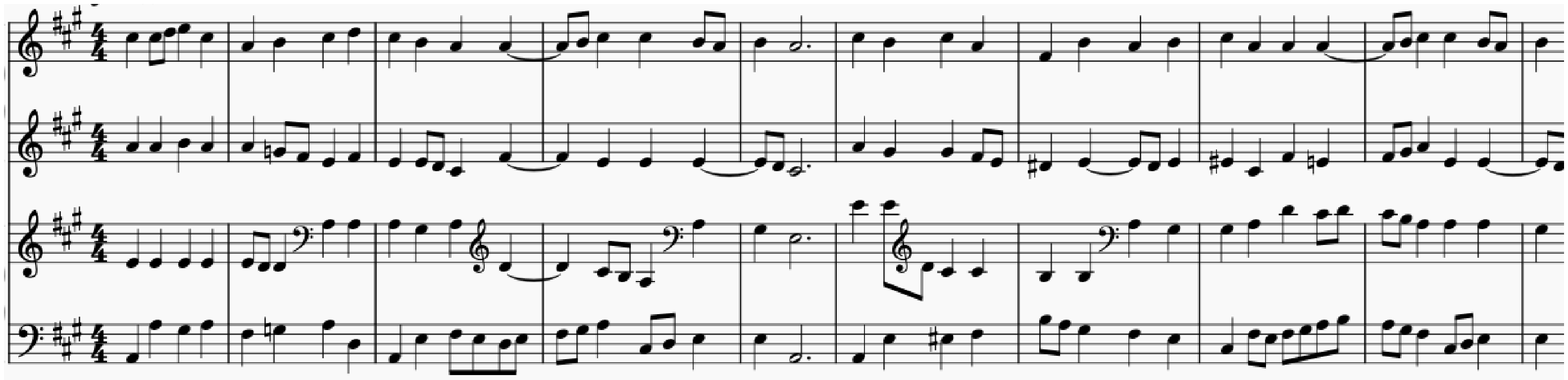}  \\ 
414 & \\
& \epsfxsize=10.cm \epsfbox{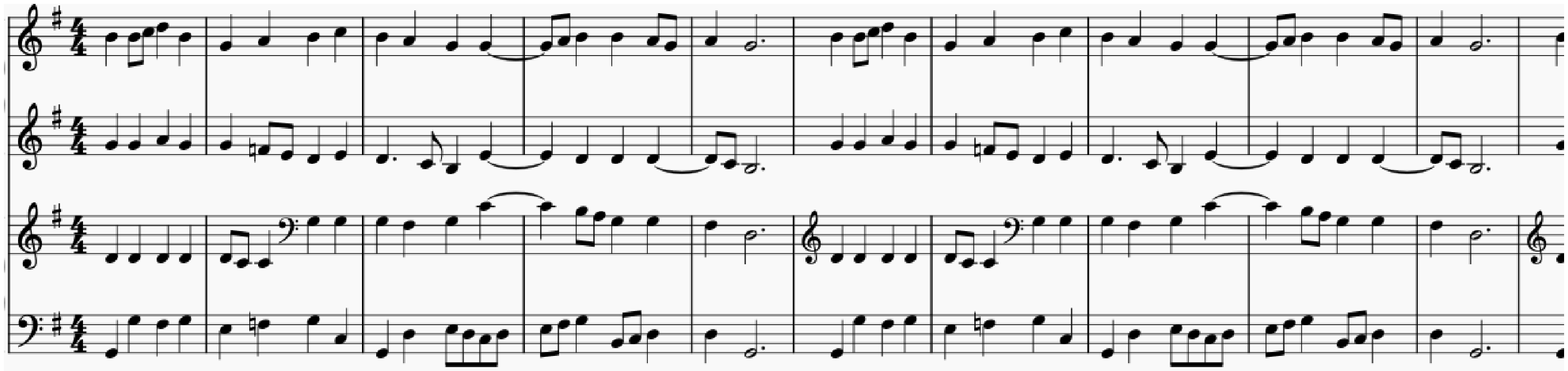}  \\ \hline \hline
318 & \\
&  \epsfxsize=10.cm \epsfbox{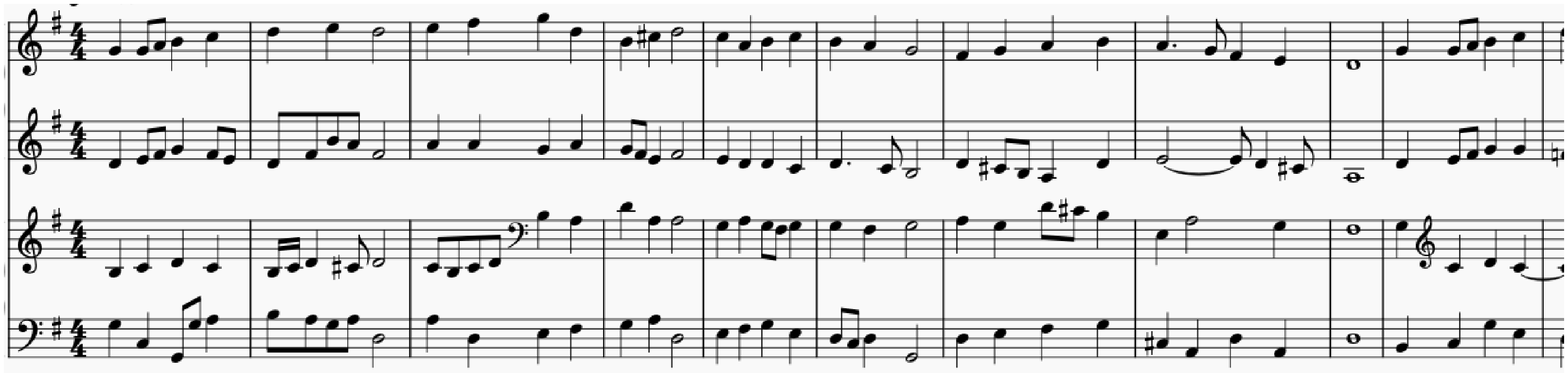}  \\  
355 &\\
& \epsfxsize=10.cm \epsfbox{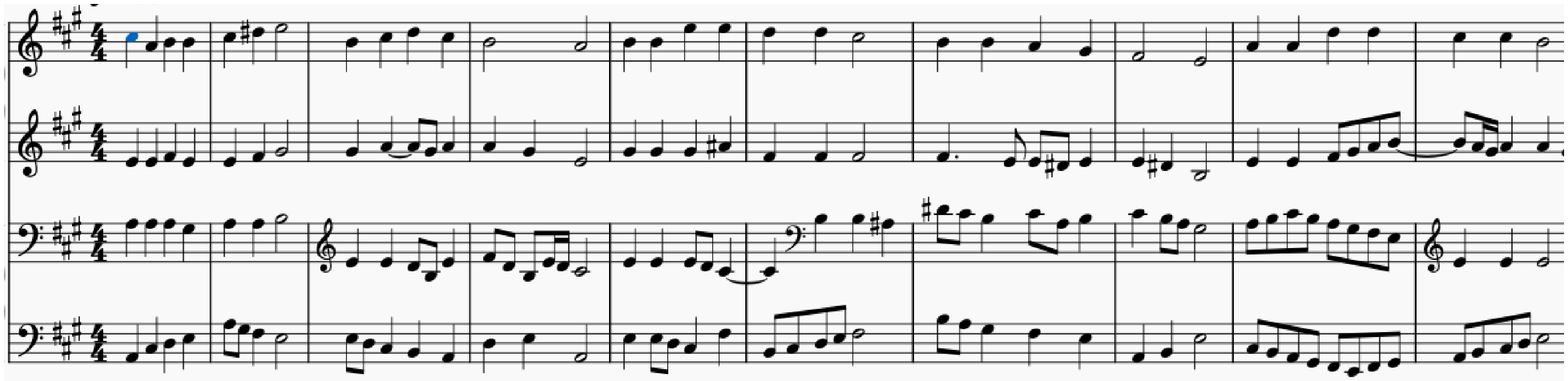}  \\ \hline \hline
\end{tabular} 
\end{small}
\end{center}
\label{harmony}
\end{table}

\subsubsection{Analyzing the Learned Node Embedding}

We evaluated the quality of learned node embedding from the three models by querying them for chorales similar to a given chorale $u$. Since  $similarity(u,v) \approx {\bf z}_u {\bf z}_v$, it was expected that ${\bf z}_v$ should represent the chorale $v$ which was similar to chorale $u$ according to the same similarity measures used for constructing the graph.

Hence, the three models i.e., skip-gram with negative sampling (SGNS), skip-gram (SG), and continuous bag-of-words (CBOW) were queried with all nodes in the graph. For each queried node, the ten most similar nodes were listed out. Common nodes between three possible model pairs SGNS-SG, SGNS-CBOW and SG-CBOW were counted. Ten-counts mean both models agreed perfectly, while zero-counts mean no matching node was found. 

We report the mean values of common nodes as well as the mean similarity values of the top ten similar nodes suggested by each model. The similarity metric  is based on the similarity of functional harmony as discussed in section \ref{choralegraph}. From Table \ref{results1}, the model SG and CBOW agree well with each other, with an average of 6.88 common nodes for every 10 nodes. The model SGNS appears to suggest a different set of nodes, with an average of 3.79 and 3.57 common nodes for SGNS-SG and SGNS-CBOW, respectively. 

Upon examining the mean similarity of the three model pairs, all pairs show comparable mean values. This means that even if SGNS suggests a different set of similar nodes (as those suggested by CBOW and SG), the similarity measures from all pairs are comparable.  Finally, three sets of similar chorales (six chorales) according to the models are shown in Figure \ref{harmony} for readers to evaluate.

\begin{table} 
\caption{Summary of similarity performances evaluated using common similar nodes between models (top three rows), and average similarity of all suggested nodes (bottom three rows).}
\begin{center}
\begin{small}
\begin{tabular}{|c|c|c|c|} \hline 
\multicolumn{3}{|c|}{Mean common nodes between} & Remarks\\
  SGNS-SG & SGNS-CBOW &  SG-CBOW &  \\ \hline
  3.79 & 3.57 & 6.88 &  p=1, q=1\\ \hline
  3.76 & 3.45 & 7.14 &  p=0.7, q=1\\ \hline
  3.67 & 3.51 & 6.94 &  p=1, q=0.7\\ \hline
\multicolumn{3}{|c|}{Mean node similarity from} & \\
  SGNS &  SG &  CBOW & \\ \hline
 8.22 & 8.82 & 8.79 & p=1, q=1\\ \hline
 8.20 & 8.80 & 8.78 & p=0.7, q=1\\ \hline
 8.27 & 8.87 & 8.78 &p=1, q=0.7\\ \hline
\end{tabular} 
\end{small}
\end{center}
\label{results1}
\end{table}

\subsection{Application Case 2: Learning Node Labels from Neighbours}
Four chorale graphs used in this task were prepared using four different arbitrary similarity threshold values. Hence the four graphs had different number of nodes, edges, and average node degrees (see Table \ref{results2a}). All edges were labeled with similarity measures according our descriptions in section \ref{choralegraph}. Each chorale node in the graph was labeled as either \emph{major} or \emph{minor} mode.

For this application case, the four graphs were initialized then their node labels were randomly removed using the following rates 10\%, 30\%, 50\%, 70\%, and 90\% from each graphs\footnote{Four different graphs and each with different missing node label rates. Hence, there are 20 experiments, see Table \ref{results2a}.}. This was to emulate the common scenario of missing labels, partially labeled dataset. We investigated the collective classification approach which aggregate information from neighboring nodes to infer the missing labels. The aggregation of information was computed using the Algorithm \ref{alg3}.

\begin{algorithm}
\caption{Collective classification}
\begin{tabbing}[!htb]
\hspace{0.1cm}  \= \hspace{0.5cm}  \= \hspace{0.5cm}  \= \hspace{0.5cm} \= \hspace{0.5cm}   \kill
\>\>\>\>\\
 Let $W$ be a weighted adjacency matrix of $\mathcal{G} = (\mathcal{V,E})$.Here weights denote\>\>\>\>\\
  \ similarity between nodes.\>\>\>\> \\
 Let label of node $u$ depends on its neighbouring labels: $P(y_u)$ = $P(y_u|N(u))$\>\>\>\>\\
 Let $y \in \{0, 0.5, 1\}$ denotes labels for class $0$, $1$ and $0.5$ denotes no label.\>\>\>\>\\
\>\>\>\> \\
 {\bf procedure} CollectiveClassification $(\mathcal{G})$ \>\>\>\>\\
\>{\bf repeat} iterations \>\>\> \\
\>\>{\bf forall} $u$ in $G$ \>\>\\
\>\>\> $ P(y_u) = \frac{1}{\sum_{(u,v) \in \mathcal{E}}W_{u,v} } \sum_{(u,v) \in \mathcal{E}} W_{u,v} P(y_v) $ \>\\
\>\>{\bf end for} \> \>\\
\>\>{\bf if} $P(y_u) >$ class 1 threshold {\bf then}  $P(y_u) = 1$ \>\>\\
\>\>{\bf if} $P(y_u) <$ class 0 threshold {\bf then}  $P(y_u) = 0$ \>\>\\
\end{tabbing}
\label{alg3}
\end{algorithm}

\begin{figure}[!ht]
\begin{center}\leavevmode
\epsfxsize=5cm
\epsfbox{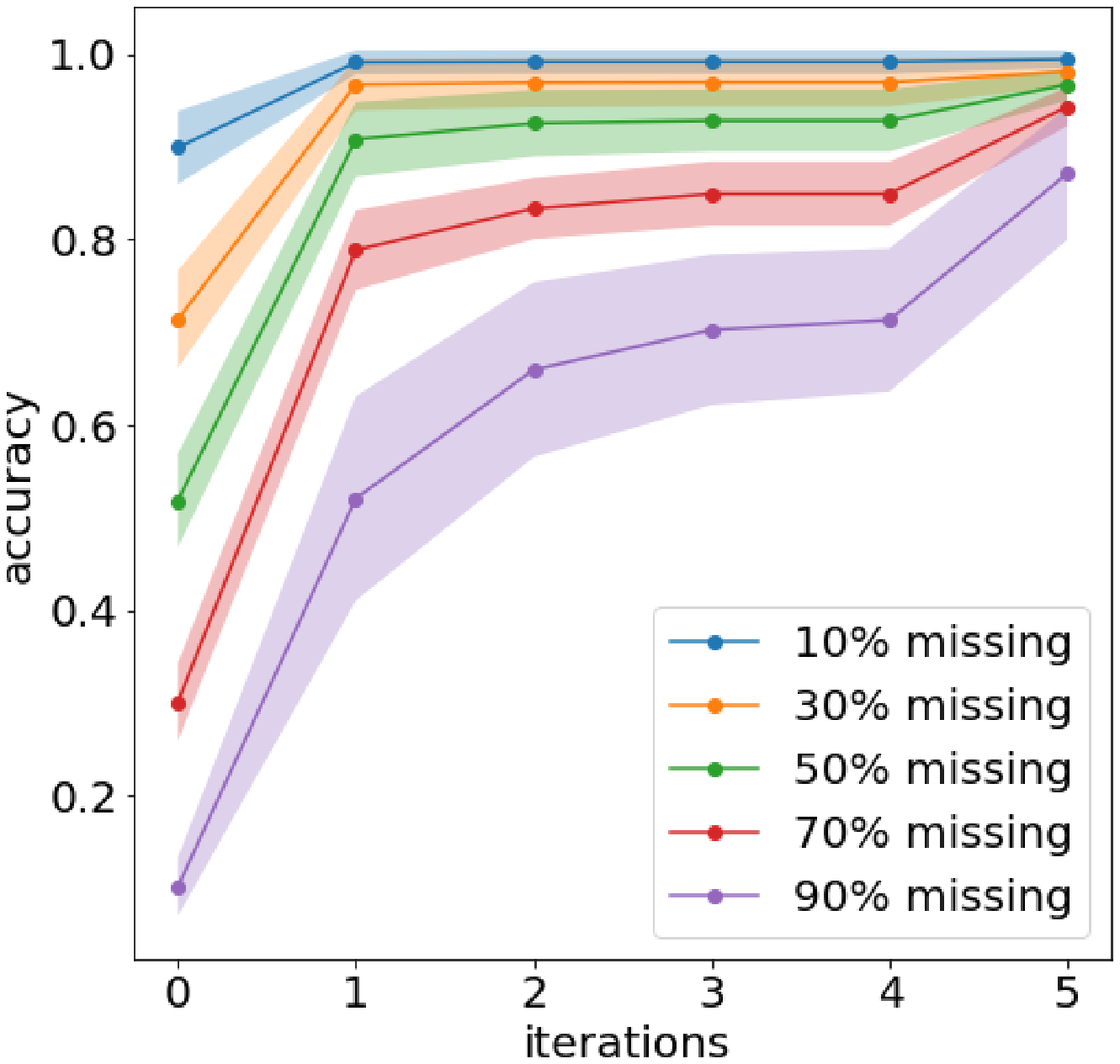}
\epsfxsize=5cm
\epsfbox{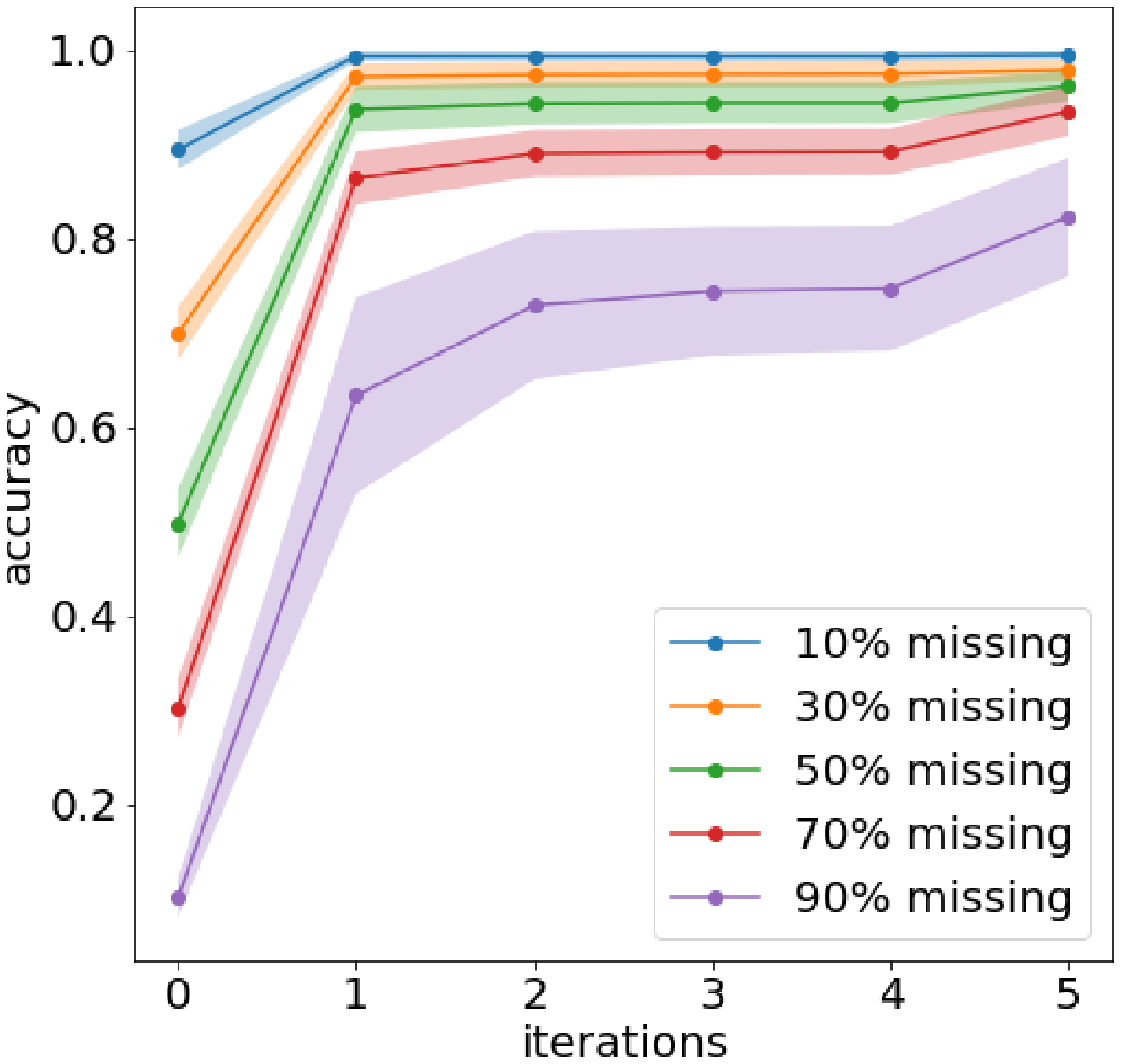}
\epsfxsize=5cm
\epsfbox{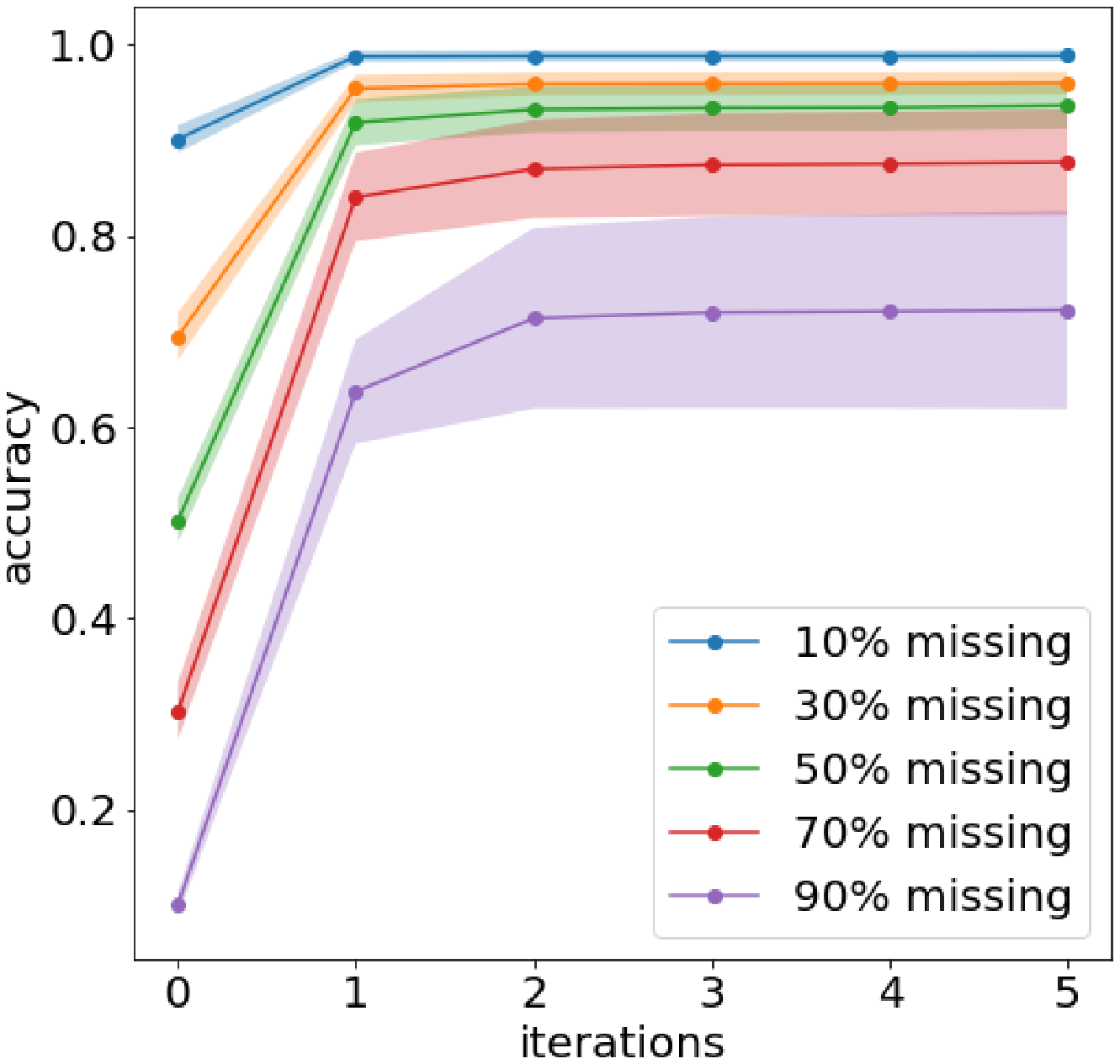}
\epsfxsize=5cm
\epsfbox{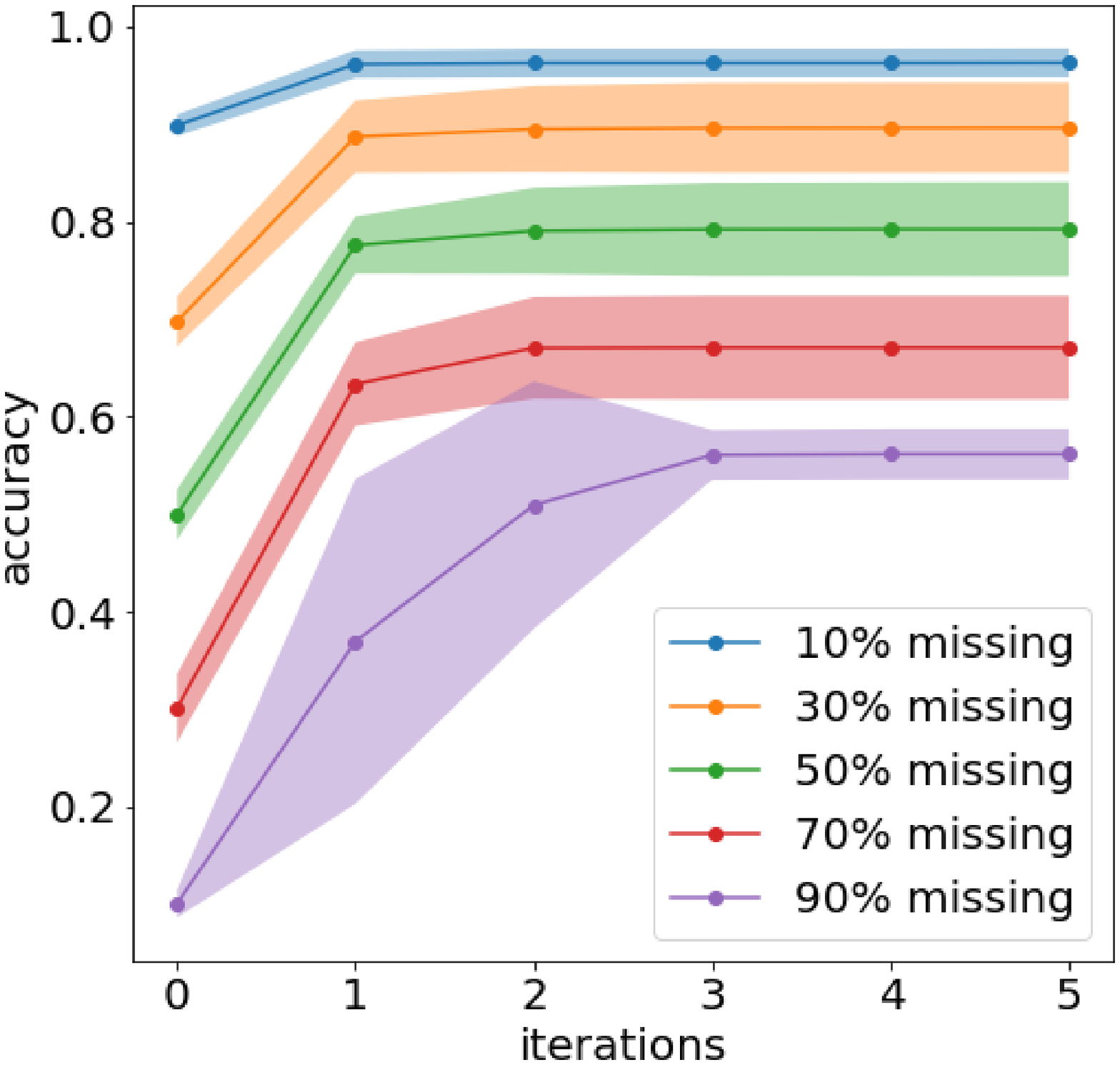}
\end{center}
\caption{Summary of results from four graphs with different rates of missing node labels after five iterations of the collective classification process.}
\label{results2b}
\end{figure}

\begin{table} 
\caption{Summary of final accuracies after five iterations. Impressive improvements are observed from small chorale graphs with lower number of nodes, edges and average node degree. This is because a smaller chorale graph is constructed with a stronger similarity constraint (in our case) and therefore reinforce the \emph{homophily} and \emph{influence} concepts in the graph network.}
\begin{center}
\begin{small}
\begin{tabular}{|c|c|c|c|c|c|c|c|} \hline 
  \multicolumn{5}{|c|}{Accuracy} &  \multicolumn{3}{c|}{Graph info.}  \\ \hline
   \multicolumn{5}{|c|}{\% missing labels} & num.&num. &avr.  \\ 
   10 &30 &50 &70 &90  &nodes &edges &degree \\ \hline  
\ 99.5\ &\ 98.1\ &\ 96.7\ &\ 94.3\ &\ 87.0\ &79 &112 &2.8 \\  
\ 99.5\ &\ 97.9\ &\ 96.1\ &\ 93.5\ &\ 82.3\ &194 &861 & 8.8\\
\ 98.8\ &\ 95.9\ &\ 93.6\ &\ 87.6\ &\ 72.2\ &365 &14556 &79.7 \\  
\ 96.3\ &\ 89.6\ &\ 79.2\ &\ 67.1\ &\ 56.2\  & 383& 73153&382 \\ \hline
\end{tabular} 
\end{small}
\end{center}
\label{results2a}
\end{table}

The underlying concepts here are \emph{homophily} and \emph{influence} in social network where characteristics of individuals within the same social group tend to correlate well.
Table \ref{results2a} summarizes the accuracy of all 20 outcomes. Each outcome is averaged from 30 runs and their standard deviation is represented by their respective shaded areas. Figure \ref{results2b} shows the plot of nodes that were correctly labeled after five iterations of message aggregation from neighbouring nodes.

\section{Conclusion \& Future Direction}
This work explores graph representation of chorales. Three hundred and eighty three Bach chorales were prepared using Music21. Each node in the graph represents a chorale composition and each edge that connects  two chorales was weighted with the similarity between them. Two application cases were explored in this study, (i) learning node embedding mapping nodes in a chorale graph to an embedded space where three algorithms were explored : node2vec, CBOW and skip-gram.; (ii) learning chorale mode labels from neighboring nodes using collective classification. 

In the first application case, the results show that node2vec (trained using negative samplings) seems to suggest a different set of similar nodes from those suggested from CBOW and SG. However, the similarity measures appear comparable. This implies that the approach is applicable to various music information retrieval tasks.
In the second application case, the missing labels can be classified correctly with a high accuracy rate. The severely missing labels case such as 90\% missing labels (10\% accuracy) could see the correct classification at 56\%-87\%. This is an increment of  46\%-77\% from 10\% correct labels at the initial stage.

In future works, we will explore various graph neural network designs for other task such as query by humming, genre classification, music synthesis, etc.

%
%


\begin{thebibliography}{}
\begin{small}
\bibitem{grover16}
Grover, A., and Leskovec, J.:
node2vec: Scalable feature learning for networks.
arXiv:1607.006533v1 (2016)

\bibitem{cuthbert10}
Cuthbert, M., Ariza, C.: 
music21: A toolkit for computer-aided musicology and symbolic music data.
In: Proceedings of the International Symposium on Music Information Retrieval, pp. 637–42, (2010) 

\bibitem{w2v13}
Mikolov, T., Chen, K., Corrado, G., Dean, J.:
Efficient estimation of word representations in vector space.
arXiv:1301.3781 (2013)

\bibitem{radimrehurek10}
Rehurek, R., Sojka, P.:
Software framework for topic modelling with large corpora.
In: Proceedings of the LREC 2010 Workshop on New Challenges for NLP Frameworks. pp. 45-50 (2010)

\bibitem{awh93}
Smaill, A., Wiggins, G. and Harris, M.:
Hierarchical music representation for composition and analysis.
Computers and the Humanities {\bf 27}(1): 7-17 (1993)

\bibitem{RPI91}
West, R. and Howell, P. and Cross, I.:
Musical structure and knowledge representation.
In: Howell, P., West, R., Cross, I. (eds.)  Representing Musical Structure, chapter  1, pp. 1-30, Academic Press, (1991)

\bibitem{courtot92}
Courtot, F.:
Logical representation and induction for computer assisted composition. 
In: M. Balaban, K. Ebcioglu, and 0. Laske, (eds.)
Understanding Music with AI: Perspectives on music cognition, chapter 7, pp. 157-181. The AAAI Press/The MIT Press. (1992)

\bibitem{span06}
Phon-Amnuaisuk, S., Smaill, A., and Wiggins, G.:
Chorale harmonisation: A view from a search control perspective.
Journal of New Music Research {\bf 35}(4): 279-305 (2006)

\bibitem{cope01}
Cope, D.: 
Virtual Music: Computer Synthesis of Musical Style. MIT Press, Oxford (2001)

\bibitem{orio09}
Orio, N. and Rod\`a, A.:
A measure of melodic similarity based on a graph representation of the music structure.
In: Proceedings of the 10th International Society for Music Information Retrieval Conference (ISMIR 2009) pp. 543-548. (2009)

\bibitem{jeong19}
Jeong, D., Kwon, T., Kim, Y., and Nam, J.:
Graph neural network for music score data and modelling expressive piano performance.
In: Proceedings of the 36th International Conference on Machine Learning. PMLR 97:3060-3070. (2019)

\bibitem{goldberg14}
Goldberg, Y., and Levy, O.:
word2vec Explained: Deriving Mikolov et al.'s negative sampling word-embedding method.
arXiv:1402.3722v1 (2014)


\end{small}
\end{thebibliography}
\end{document}